\pdfoutput=1
\documentclass[11pt]{article}

\usepackage[]{acl}
\usepackage{times}
\usepackage{latexsym}
\usepackage{pifont}
\usepackage[T1]{fontenc}
\usepackage[utf8]{inputenc}
\usepackage{microtype}
\usepackage{inconsolata}
\usepackage{graphicx}
\usepackage{amsmath}
\usepackage{amsthm}
\usepackage{amsfonts}
\usepackage{bm}
\usepackage{caption}
\usepackage{multirow}
\usepackage[english]{babel}
\usepackage{subfigure}
\usepackage{booktabs}
\usepackage{array}
\usepackage{algorithm}
\usepackage{algorithmic}
\theoremstyle{definition}
\newcommand{\model}{\textsc{BioRAG}}

\title{\model: A RAG-LLM Framework for Biological Question Reasoning}

\author{
  Chengrui Wang\textsuperscript{\rm 1,2}, Qingqing Long\textsuperscript{\rm 1,2}, Meng Xiao\textsuperscript{\rm 1,2}, Xunxin Cai\textsuperscript{\rm 1,2},
  Chengjun Wu\textsuperscript{\rm 1,2}, \\ \textbf{Zhen Meng}\textsuperscript{\rm 1,2}, \textbf{Xuezhi Wang}\textsuperscript{\rm 1,2}, \textbf{Yuanchun Zhou}\textsuperscript{\rm 1,2}\thanks{Yuanchun Zhou is the corresponding author.}
  \\
  \textsuperscript{\rm 1}Computer Network Information Center, Chinese Academy of Sciences. \\ 
  \textsuperscript{\rm 2}University of the Chinese Academy of Sciences.\\
    \texttt{\{crwang,qqlong,shaow,xxcai,cwu,zhenm99,wxz,zyc\}cnic.cn}\\
}

\begin{document}
\maketitle
\begin{abstract}
The question-answering system for Life science research, which is characterized by the rapid pace of discovery, evolving insights, and complex interactions among knowledge entities, presents unique challenges in maintaining a comprehensive knowledge warehouse and accurate information retrieval. 
To address these issues, we introduce \textbf{\model}, a novel Retrieval-Augmented Generation (RAG) with the Large Language Models (LLMs) framework.
Our approach starts with parsing, indexing, and segmenting an extensive collection of 22 million scientific papers as the basic knowledge, followed by training a specialized embedding model tailored to this domain. 
Additionally, we enhance the vector retrieval process by incorporating a domain-specific knowledge hierarchy, which aids in modeling the intricate interrelationships among each query and context.
For queries requiring the most current information, \model\ deconstructs the question and employs an iterative retrieval process incorporated with the search engine for step-by-step reasoning. 
Rigorous experiments have demonstrated that our model outperforms fine-tuned LLM, LLM with search engines, and other scientific RAG frameworks across multiple life science question-answering tasks.
\end{abstract}

\section{Introduction}
Research and trends in the \textit{Biology} have shown a continuously evolving, marked by rapid discoveries and the increasing complexity of its knowledge domains~\cite{bertoline2023before,long2021hgk}. 
In addition, the growing trend for interdisciplinary research between Biology and other fields~\cite{lepore2023interdisciplinary,xiao2023hierarchical,xiao2023interdisciplinary}, such as artificial intelligence~\cite{holzinger2023ai,long2021theoretically}, material science~\cite{atkins2023physical}, and environmental science~\cite{cole2021plant}, further amplifies the complexity of knowledge synthesis. 
To bridge the gap and facilitate multidiscipline cooperation, automated question-reasoning systems~\cite{auer2023sciqa} play a pivotal role in enabling experts from diverse fields to effectively navigate and integrate this burgeoning and complex body of biological knowledge~\cite{yang2023empower}.
However, this ever-changing landscape and the complex interplay between different knowledge components present obstacles~\cite{lee2023benefits,castro2023large,lecler2023revolutionizing,song2020deep} in creating efficient domain-specific question-reasoning systems. 

\begin{figure}[t]
  \includegraphics[width=\linewidth]{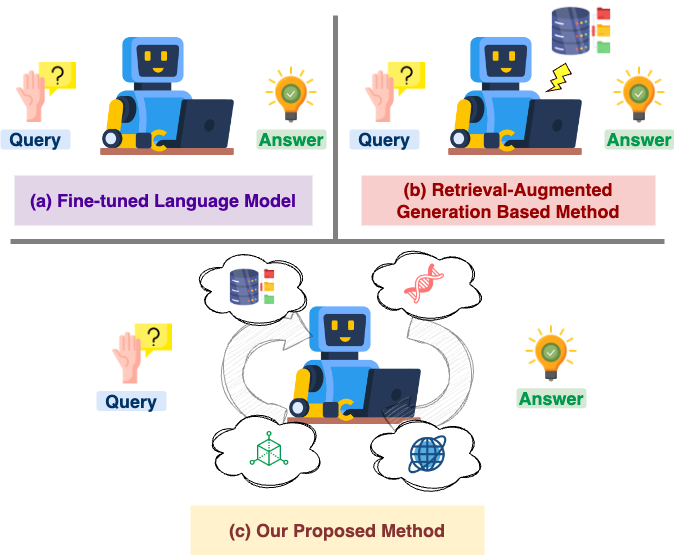}
  \caption {An illustration of the difference between three paradigms: (a) fine-tuned language model embedded domain knowledge into deep space; (b) RAG-based method retrieve supplementary information from constructed knowledge base; (c) \model\ adaptively select knowledge source and domain-specific tools to advance the biology question-reasoning task.}
\label{motivation}
\end{figure}

The prior literature partially addresses question-reasoning in the biology domain and can be grouped into two mainstream~\cite{nguyen2024enhancing} (as shown in Figure~\ref{motivation} (a-b)). 
\textbf{Fine-tuned Language Model}~\cite{gu2021domain} includes models like bioBERT~\cite{lee2020biobert}, sciBERT~\cite{beltagy2019scibert}, and large language models tailored for specific domains, such as PMC-Llama~\cite{wu2024pmc} and Llava-med~\cite{li2024llava}.
These models are trained on domain-specific corpora, thereby embedding deep domain knowledge within their architectures. 
\textit{However, that embedded knowledge could be incomplete and computationally expensive to update.}
\textbf{Retrieval-Agumented Generation} methods follow the information indexing and retrieval, information augmentation, and answer generation paradigm. 
For instance, PGRA~\cite{guo2023prompt} adopts a retriever to search and re-ranking the context, then generate the answer.
Later research has aimed to improve these systems by either optimizing the retrieval processes using prior answers~\cite{wang2023self}, enhancing model functionality through iterative feedback cycles~\cite{liu2024ra}, or expanding the knowledge base with search engines to incorporate the latest information~\cite{o2023new}. 
Although RAG-based methods address the issue of updating information, \textit{they often oversee the intricate complexities inherent in the domain knowledge of biology.}

\begin{figure*}[t]
    \centering
  \includegraphics[width=\linewidth]{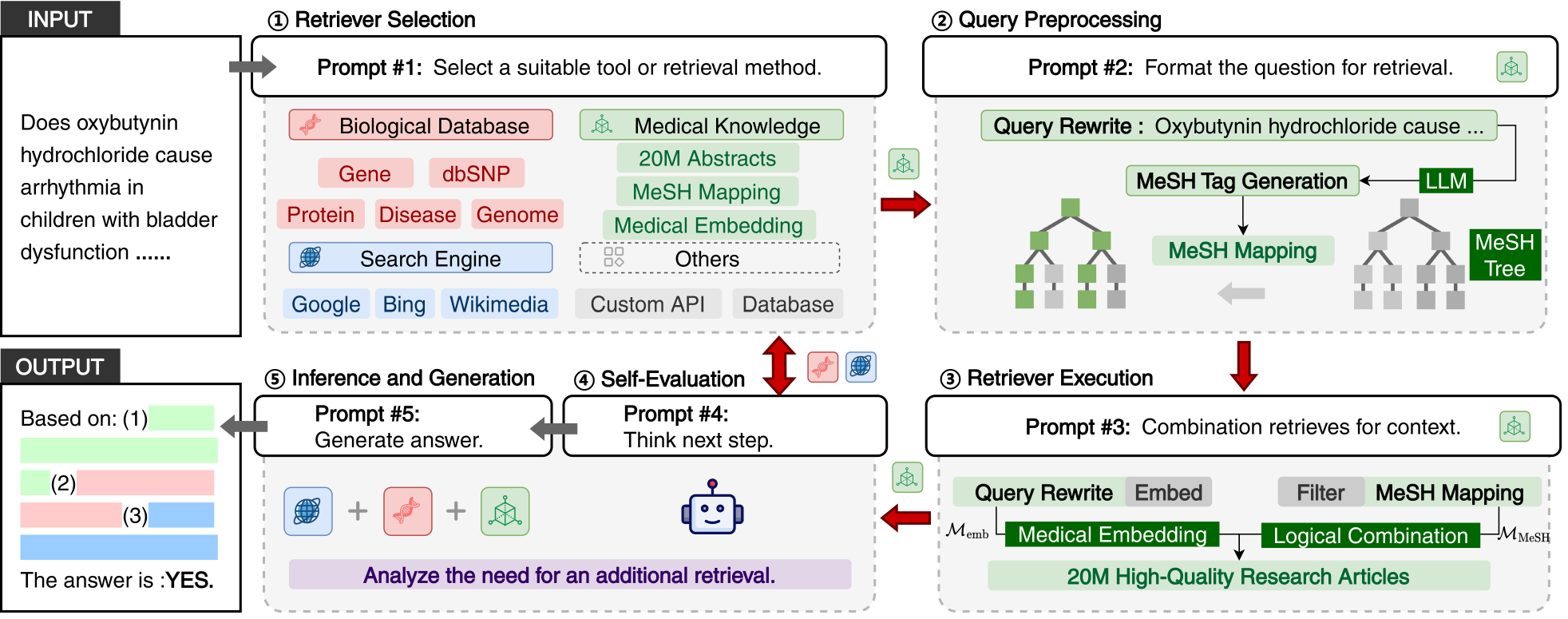}
\caption {The architecture of our proposed \model\ framework. The pipeline consists of five iterative components designed to enhance the process of biological question-reasoning: \textbf{\ding{172} Retriever Selection} aims to choose the most ideal information source; \textbf{\ding{173} Query Pre-processing} aims to rewrite the query and find closed topic tag from pre-defined knowledge hierarchy; \textbf{\ding{174} Retriever Execution} aims to combination retrieve the correlated context from knowledge base; \textbf{\ding{175} Self-Evaluation} assess the adequacy of the retrieved information and decides whether to cycle through additional retrieval tools or to move to the next phase; \textbf{\ding{176} Inference and Generation} uses the information gathered to generate an informed and accurate answer to the biological query.}
  \label{fig:framework}
\end{figure*}

Based on the aforementioned discussion, we summarize three challenges in building efficient biology question-reasoning systems: 
\textbf{(C1) The scarcity of high-quality domain-specific corpora.} 
While biological research publications are abundant, there remains a significant void in the availability of extensive, high-quality datasets to build robust information indexing models.
\textbf{(C2) The inherent complexity of biological knowledge systems.} 
This complexity is compounded by the interdisciplinary nature of modern biological research. 
Consequently, automated question-reasoning systems must be able to understand and process multifaceted and often ambiguous biological query.
\textbf{(C3) The continual updating of knowledge.}
Biology is a dynamic field where discoveries are frequently made, and existing theories are regularly revised or replaced. 
This fluidity necessitates that question-reasoning systems adeptly select the knowledge source from databases or contemporary search engines to reflect the correct scientific understanding. 

\textbf{Our Perspective and Contributions:} 
To solve the above challenges, we proposed \model, a novel Retrieval-Augmented Generation framework integrated with Large Language Models for biological question-reasoning. 
To obtain a robust domain-specific information indexing embedding model, we start by parsing, indexing, and segmenting extensive research articles from the biology domain and constructing high-quality training corpora. 
\model\ then addresses the complexity of biological knowledge systems by combining a pre-built research hierarchy with an embedding model for accurate context retrieval. 
To cope with emerging biology knowledge, \model\ can adaptively select knowledge sources from search engines, existing domain-specific tools, or indexed research articles.
Once the framework determines that it has gathered sufficient information, it will generate the answer based on the reasoned material.

We illustrate the question-reasoning power of \model\ on 6 popularly used biology QA datasets and compare it against 6 baseline methods. 
Extensive case studies show the great potential to apply this framework to general science question-reasoning scenarios.

\section{Biological Retrieval-Augmented Generation LLM Framework}

In this paper, we propose the \textit{\textbf{Bio}logical \textbf{R}etrieval-\textbf{A}ugmented \textbf{G}eneration LLM Framework}, namely \model\ (as shown in  Figure~ \ref{fig:framework}). 
In the following sections, we first introduce the preliminary step of constructing a high-quality local information source and training the biological domain-specific information indexing embedding model.
For questions that require the most current or other domain-related data, we introduce external information sources. 
Then, we demonstrate the knowledge hierarchy-based query pre-processing, retriever execution component, and how the model iteratively collects sufficient information. 
Finally, the large language model will generate the answer based on the information obtained. 
The details of customized prompts are given in Section~\ref{prpt}.

\subsection{Internal Biological Information Source}
High-quality domain-specific corpora are crucial for enriching the information source and enhancing the embedding model in the context of biological question-reasoning systems. 
To achieve this goal, we extract research papers from the global biomedical article database maintained by the National Center for Biotechnology Information\footnote{\url{https://www.ncbi.nlm.nih.gov/}}(NCBI)~\cite{schoch2020ncbi}. 
This extensive repository aggregates over 37 million scientific citations and abstracts spanning from the 1950s to the present, encompassing a broad array of biomedical fields, including clinical medicine, molecular biology, etc.
For the purposes of this study, we utilize the abstracts from these PubMed papers as the supporting corpus for the \model\ framework.

\smallskip
\noindent\underline{\textit{Local Data Preparation:}}  Specifically, we initially downloaded over 37 million original papers from which we subsequently filtered out 14 million entries deemed to be of low quality. The preprocessing of these texts was conducted using the \textit{Unstructured} tool\footnote{\url{https://github.com/Unstructured-IO}}, specifically designed to ingest and preprocess unstructured textual data effectively. 
Our filtration process involved the removal of gibberish using regular expression techniques, as well as the exclusion of non-semantic content such as hyperlinks, charts, tables, and other embedded tags. This meticulous process yielded a corpus of 22,371,343 high-quality, processed PubMed abstracts.

\smallskip
\noindent\underline{\textit{Information Indexing:}} To further refine the retrieval performance of abstracts tailored to specific biological questions, we developed a specialized biological embedding model within the \model\ framework.
This model employs PubMedBERT~\cite{gu2021domain} as the foundational model.
We enhanced this model using the CLIP (Contrastive Language-Image Pretraining) technique \citep{li2021supervision, nomic}, allowing us to fine-tune the model, denoted as $\mathcal{M}_{\text{emb}}$.
Based on this, we constructed a local, high-quality biological vector database \citep{xian2024vector} to support efficient and effective query processing and retrieval operations. This database serves as a critical resource in facilitating rapid and accurate access to relevant biomedical information, significantly advancing the capabilities of our \model framework in handling complex biological questions.

\subsection{External Information Sources}
External biology knowledge is crucial to biological reasoning due to the rapidly evolving nature of biological research, which continuously integrates new discoveries. 
To address this challenge, we introduce two external information sources.

\smallskip
\noindent\underline{\textit{Biological Data Hub:}} 
In \model, we harness several specialized biological Hubs to ensure the accuracy of experimental data and to provide detailed biological insights. 
Specifically, \model integrates the following databases, each serving a unique purpose in the broader context of biological analyses:
\noindent{(1) Gene Database}\footnote{\url{https://www.ncbi.nlm.nih.gov/gene/}}: This resource provides comprehensive information on the functions, structures, and expressions of specific genes. It is invaluable for addressing queries related to gene mechanisms, gene actions, and gene expressions, facilitating a deeper understanding of gene-related phenomena.
\noindent{(2) dbSNP Database}\footnote{\url{https://www.ncbi.nlm.nih.gov/snp/}}: This database houses a vast repository of single nucleotide polymorphisms (SNPs), offering critical insights into genetic variants and their potential associations with various diseases. It is instrumental for studies exploring the genetic basis of disease and trait inheritance.
\noindent{(3) Genome Database}\footnote{\url{https://www.ncbi.nlm.nih.gov/genome/}}: Providing complete genome sequences, this database is essential for studying the structure, function, and evolution of genomes across different organisms. It supports comprehensive genomic analyses and comparative studies, enhancing our understanding of genomic architecture and its functional implications.
\noindent{(4) Protein Database}\footnote{\url{https://www.ncbi.nlm.nih.gov/protein/}}: This resource offers detailed information about the sequences, structures, and functions of proteins. It is crucial for exploring protein-related biological processes, understanding molecular functions, and investigating the complex interactions within the proteome.

\smallskip
\noindent\underline{\textit{Search Engine:}} To ensure access to the most current discussions and developments, \model\ incorporates a variety of search engines, including Google, Bing, arXiv, Wikimedia, and Crossref. Each platform contributes uniquely to the aggregation of information:
(1) Google and Bing: These search engines scour the web for a diverse range of content, including news articles, blogs, and forums, providing insights into public discussions and concerns related to scientific topics. This breadth of information is crucial for understanding the societal impact and general discourse surrounding scientific issues.
(2) arXiv: As a repository for preprint papers, arXiv offers access to the latest research reports and scholarly articles across multiple scientific disciplines before they undergo peer review. This source is invaluable for staying abreast of the newest scientific theories and experiments.
(3) Wikimedia: Known for its user-friendly content, Wikimedia offers easily digestible explanations of complex scientific concepts and principles. This resource helps simplify advanced topics for broader public understanding and educational purposes.
(4) Crossref: This service acts as a comprehensive aggregator of academic citation data, providing links to peer-reviewed scholarly publications and their citation networks. Crossref is essential for accessing high-quality research outputs and understanding their impact on the academic community.

\begin{figure}[!ht]
    \framebox{
    \parbox{0.45\textwidth}{
    \small
    Based on the \textbf{QUESTION}, analyze the related MeSH terms to format them properly.  \newline
    \textbf{QUESTION}: [.....]  \newline
    \textbf{MeSH}: [$\kappa_1$, $\kappa_2$, ...]
    }
    }
    \caption{Training Template for $\mathcal{M}_{\text{MeSH}}$.} 
    \label{fig:train_mesh}
\end{figure}

\begin{figure}[!ht]
    \framebox{
    \parbox{0.45\textwidth}{
    \small
    \colorbox{gray!20}{\textbf{Input Question}} \newline
    What are the differences between innate immunity and adaptive immunity?\newline \newline
    \colorbox{red!20}{\textbf{Predicted MeSH by $\mathcal{M}_{\text{MeSH}}$}} \newline
    [Adaptive Immunity, Animals, ...]\newline \newline
    \colorbox{green!20}{\textbf{Generated SQL}} \newline
    {"\textbf{filtered by}":[eq("MeSH", "Adaptive Immunity") or eq("MeSH", "Animals") or ...], \newline
    "\textbf{ordered by}": embedding similarity}
    }
    }
    \caption{An example of MeSH filtering SQLs Generation. } 
    \label{fig:self_query}
\end{figure}

\begin{table*}
\centering
\scalebox{0.8}{
\begin{tabular}{lccccccccc}
\toprule
\multirow{2}{*}{} & \multicolumn{3}{c}{\textbf{LLM}} & \multicolumn{2}{c}{\textbf{BioLLM}} & \multicolumn{2}{c}{\textbf{SciRAG}} & \multirow{2}{*}{\textbf{BioRAG}} \\
\cmidrule(lr){2-4} \cmidrule(lr){5-6}  \cmidrule(lr){7-8}  
& GPT3.5 & Llama3-8B &  Llama-70B & PMC-Llama & BioMistral & GeneGPT & NewBing  & \\
\midrule
\multicolumn{7}{l}{\textbf{Nomenclature}}\\ \midrule
Gene alias & 7 & 0 & 0 & 0 & 8 & \underline{84} & 68 &\textbf{98} \\
Gene name conversion & 0 & 0 & 0 & 0 & 0 & \textbf{100} & \textbf{100} & \textbf{100} \\
\midrule
\multicolumn{7}{l}{\textbf{Genomic location}} \\ \midrule
Gene SNP association & 0 & 0 & 0 & 0 &  0 & \textbf{100} & 0 &  \textbf{100} \\
Gene location & 9 & 20 & 28 & 14 & 12 &  66 & \underline{70} & \textbf{86} \\
SNP location & 5 & 48 & 94 & 0 & 0 & 98 & \textbf{100} & \textbf{100} \\
\midrule
\multicolumn{7}{l}{\textbf{Functional analysis}} \\ \midrule
Gene disease association & 31 & 0 & 0 & 0 & 8 & \underline{66} & 64 & \textbf{71} \\
Protein-coding genes & 54 & 6 & 12 & 40 & \underline{80} & \textbf{100} & \textbf{100} & \textbf{100} \\
\bottomrule
\end{tabular}
}
\caption{Performance of BioRAG compared to other RAG-LLMs on the GeneTuring QA dataset.The scores represent accuracy.}
\label{tab:res_turing}
\end{table*}

\subsection{Self-evaluated Information Retriever}
Following the construction of the internal and external information source, \model\ is firstly tasked with comprehending the complex disciplinary framework of the life sciences to retrieve the most relevant information accurately.
Moreover, \model\ integrates a self-evaluation mechanism to continuously assess the adequacy and relevance of the information it has collected.

\smallskip
\noindent\underline{\textit{Internal Information Retrieve: }} To effectively navigate the inherent complexity of biological knowledge systems, \model\ leverages an integrated approach, combining a well-defined hierarchical structure with indexed information to conduct a comprehensive internal information retrieval.
The Medical Subject Headings\footnote{\url{https://www.nlm.nih.gov/mesh/meshhome.html}} (MeSH) thesaurus is popularly used for indexing, cataloging, and searching for biomedical-related information and research papers. 
Specifically, we first train a model $\mathcal{M}_{\text{MeSH}}$ to predict MeSH of the input questions. 
We then use the templates in Figure~\ref{fig:train_mesh} for fine-tuning a Llama3-8B model to classify given questions.
After that, we construct MeSH filtering SQLs (as shown in Figure~\ref{fig:self_query}) to generate the scalar condition retrieval.
A candidate result is considered relevant to the given question because it has one consistent MeSH with the question. 
Then, the vector retrieval process is adopted to sort the relative results based on the cosine similarity of the sentence embedding between the input questions and the filtered results.

\begin{table*}
\centering
\resizebox{\linewidth}{!}{
\begin{tabular}{lcccccccccccc}
\toprule
\multirow{2}{*}{} & \multicolumn{3}{c}{\textbf{LLM}} & \multicolumn{2}{c}{\textbf{BioLLM}} & \multicolumn{2}{c}{\textbf{SciRAG}} & \multirow{2}{*}{\textbf{BioRAG} } \\
\cmidrule(lr){2-4} \cmidrule(lr){5-6}  \cmidrule(lr){7-8}  
& GPT3.5 & Llama3-8B &  Llama-70B & PMC-Llama & BioMistral & GeneGPT & NewBing  & \\
\midrule
MedMCQA & 54 & 51 &  \underline{71} & 56 & 49 & 0 & 55 & \textbf{73} \\
Medical Genetics &  \underline{74} & 51 & 67 & 28 & 67 & 0 & \textbf{88} & \textbf{88} \\
College Biology & 73 & 75 &  \underline{88} & 30 & 67 & 0 & 71 & \textbf{90} \\
College Medicine & 65 & 61 & \underline{70} & 23 & 51 & 0  & \textbf{78}  & \textbf{78} \\
\bottomrule
\end{tabular}}
\caption{Performance of BioRAG compared to other RAG-LLMs on the biological-related QA benchmarks.The scores represent accuracy. \textbf{Bold} and \underline{underlined} results denote the highest and second-highest performance, respectively. }
\label{tab:res}
\end{table*}

\smallskip
\noindent\underline{\textit{Self-evaluation Strategy:}} 
In order to ensure the accuracy and contemporary of the retrieved information, \model\ incorporates a self-evaluation strategy that assesses the adequacy of data collected from the internal knowledge base.
In detail, this critical evaluation is driven by the backend large language model which aims to determine whether the information retrieved internally is sufficient to address the posed question substantively. 
If the internal content is insufficient, the model will loop back to pertinent external knowledge sources. 
Additionally, when the initial assessment indicates that the scientific questions require broader searches or retrieval of entity-specific data, the model tends to deploy external tools.
This methodology supports the framework's goal of providing precise, up-to-date, comprehensive answers, facilitating more informed decision-making, and advancing research and applications in the life sciences.

\subsection{Customized Prompts Detail}\label{prpt}
To maximize the effect of the retrieved corpus and knowledge, we design customized prompts in \model. The prompts in Figure. \ref{fig:framework} is detailed defined as follows,
\begin{itemize}
    \item \textbf{Prompt \# 1}: To provide the most helpful and accurate response to the following Question: \textit{\{Question\}}. You have been given descriptions of several RETRIEVAL METHODS: \textit{\{Retrieval\}}. Please select the RETRIEVAL METHODS you consider the most appropriate for addressing this question.
    \item \textbf{Prompt \# 2}: Based on the RETRIEVAL METHODS you selected, and considering the \textit{Question} and the \textit{Input Requirements} of the retrieval method, please REWRITE the search query accordingly.
    \item \textbf{Prompt \# 3}: Now, using the rewritten QUERY and the retrieval FILTER methods, perform a logical combination to execute the search effectively.
    \item \textbf{Prompt \# 4}: Based on the RETRIEVAL RESULTS from the above steps, please evaluate whether the RESULTS support answering the original \textit{Question}. If they do not support it, output "\textbf{NO}". If they do support it, output "\textbf{YES}".
    \item \textbf{Prompt \# 5}: Based on the RETRIEVAL RESULTS, perform a comprehensive reasoning and provide an answer to the \textit{Question}.
\end{itemize}

Furthermore, we designed instruction manuals for specialized biological tools and databases, aim at exploiting their potentialities. These instructions are shown as follows,
\begin{itemize}
    \item \textbf{Manual \#Gene}: The Gene database search engine is a valuable tool for retrieving comprehensive information about genes, including gene structure, function, and related genetic events. It is particularly useful for answering detailed questions regarding gene-related research and findings.  To utilize this search engine effectively, the input must be a specific gene name.
    \item \textbf{Manual \#dbSNP}: The dbSNP database search engine is an essential tool for retrieving detailed information about single nucleotide polymorphisms (SNPs) and other genetic variations. It is particularly useful for answering questions related to genetic diversity, allele frequency, and related genetic studies. To utilize this search engine effectively, the input must be a specific SNP identifier or genetic variant name.
    \item \textbf{Manual \#Genome}: The Genome database search engine is an indispensable tool for accessing comprehensive information about entire genomes, including their sequences, annotations, and functional elements. It is particularly useful for answering complex questions about genomic structures, variations, and comparative genomics. To use this search engine effectively, the input must be a specific genome name or identifier.
    \item \textbf{Manual \#Protein}: The Protein database search engine is a crucial resource for obtaining detailed information about proteins, including their sequences, structures, functions, and interactions. It is particularly useful for answering questions related to protein biology, biochemical properties, and molecular function. To use this search engine effectively, the input must be a specific protein name or identifier.
    \item \textbf{Manual \#Web Search}: The Web Search Engine is a powerful tool designed to help you find information about current events quickly and efficiently. It is especially useful for obtaining the latest news, updates, and developments on a wide range of topics. To use this search engine effectively, simply enter a relevant search query.
    \item \textbf{Manual \#PubMed}: The PubMed local vector database search engine is an advanced tool designed for retrieving biomedical literature and research articles using vector-based search techniques. It is particularly useful for answering detailed questions about medical research, clinical studies, and scientific discoveries. To utilize this search engine effectively, the input should be a specific query or topic of interest.
\end{itemize}

\section{Results \& Analysis}
\subsection{Datasets}
We conduct experiments on 6 popularly used biological-related QA datasets to evaluate our proposed \model, i.e.,  GeneTuring \citep{hou2023geneturing}, MedMCQA \cite{pal2022medmcqa}, Medical Genetics \cite{hendrycks2020measuring}, College Biology \cite{hendrycks2020measuring},  College Medicine  \cite{hendrycks2020measuring}. Note that the GeneTuring dataset contains more specialized biological questions. It contains 12 tasks, and each task has 50 question-answer pairs. We use 7 GeneTuring tasks that are related to NCBI resources to evaluate the proposed \model. The chosen tasks are classified into three modules and briefly described as follows,
\begin{itemize}
    \item \textbf{Nomenclature:} This is about gene names. The objectives of the gene alias task and name conversion task are finding the official gene symbols for their non-official synonyms.
    \item \textbf{Genomics location:} The tasks are about the locations of genes, single-nucleotide polymorphism (SNP), and their relations. We include the gene location, SNP location, and gene SNP association tasks. The first two tasks ask for the chromosome locations of a gene or an SNP, and the last one asks for related genes for a given SNP.
    \item \textbf{Functional analysis} asks for gene functions. We use the gene-disease association task where the goal is to return related genes for a given disease, and the protein-coding genes task which asks whether a gene is a protein-coding gene or not.
\end{itemize}
\begin{table*}
\centering
\scalebox{0.85}{
\begin{tabular}{lccccccccccc}
\toprule
\multirow{2}{*}{} & \multicolumn{3}{c}{\textbf{Data}} & \multicolumn{3}{c}{\textbf{Component}} & \multicolumn{2}{c}{\textbf{Base Model}} &  \\
\cmidrule(lr){2-4} \cmidrule(lr){5-7}  \cmidrule(lr){8-9}  

& D1 & D2 & D3 & C1 & C2 & C3 & M1 & M2 \\
\midrule
\multicolumn{7}{l}{\textbf{Nomenclature}}\\ \midrule
Gene\_location & 74 & 94 & 96 & 98 & 90 & 91 & 88 & 98 \\
SNP\_location & 50 & 100 & 100 & 100 & 100 & 100 & 92 & 100 \\ \midrule
\multicolumn{7}{l}{\textbf{Genomic location}} \\ \midrule
Gene\_SNP\_association & 6 & 100 & 98 & 100 & 6 & 100 & 94 & 100 \\
Gene\_disease\_association & 82 & 60 & 84 & 84 & 84 & 36 & 70 & 86 \\
Protein\_coding\_genes & 20 & 100 & 100 & 100 & 18 & 100 & 100 & 100 \\ \midrule
\multicolumn{7}{l}{\textbf{Functional analysis}} \\ \midrule
Gene\_name\_conversion & 64 & 70 & 62 & 70 & 66 & 32 & 64 & 71 \\
Gene\_alias & 100 & 100 & 100 & 100 & 98 & 80 & 92 & 100 \\
\bottomrule
\end{tabular}
}
 \caption{Ablation study on the GeneTuring dataset.The scores represent accuracy.}
\label{tab:ablation}
\end{table*}

\subsection{Baslines}
We compare \model  with various baselines, which can be classified into three categories,
\begin{itemize}
    \item \textbf{LLM} (General LLMs): We select GPT-3.5-Turbo (175B), Llama3-8B (8B), Llama-70B (70B) as representative baselines.
    \item \textbf{BioLLM} (Biological LLMs): PMC-Llama (13B) \citep{wu2024pmc} and BioMistral (7B) \citep{labrak2024biomistral} are two medical LLMs. They are pre-trained on open-source biomedical texts. 
    \item \textbf{SciRAG} (Scientific RAG-LLM framework): GeneGPT (175B) \cite{genegpt} is a biological RAG-LLM framework that integrates the NCBI databases, i.e., Gene, dbSNP, OIMI, and Blast. NewBing \footnote{\url{https://www.bing.com/chat}} (>400B)  is a retrieval-augmented LLM that has access to relevant web pages retrieved by Bing. 
\end{itemize}

\subsection{Experimental Settings}
We take the Llama3-70B as the basic language model of \model. For our embedding model $\mathcal{M}_{\text{emb}}$, we take AdamW as the optimizer and fine-tune 2 epochs. 
The number of retrieved results by biological databases, search engines, and local PubMed databases are set to 10, 10, and 4, respectively. The max iteration of self-evaluation is set to 15. If the model does not output the final answer within 15 times, \model\ stops the iteration and outputs the current wrong answer. We use the \textit{accuracy} to verify the overall performance. For the GeneTuring dataset, we only consider \textit{exact} matches between model predictions and the ground truth as correct predictions for all nomenclature and genomics location tasks. For the gene-disease association task, we measure the recall as in the original dataset but based on \textit{exact} individual gene matches.
For the protein-coding genes task, we consider \textit{exact} matches as correct after applying a simple vocabulary mapping that converts model-predicted "yes" / "no" to "TRUE" / "NA" and Latin species names to their informal names, respectively. The final answer of other datasets is "yes" / "no".

\subsection{Results on Biological-related Tasks}
To verify the effectiveness of the proposed model, we first conduct biological QA tasks. Results are shown in Table~\ref{tab:res}. We conclude with the following findings:
(1) Based on the results of BioLLMs and GPT-3.5, we conclude that fine-tuning domain-specific data is helpful for domain-specific tasks. As the size of BioLLMs is much smaller than GPT-3.5, their performance is on par with GPT-3.5. 
(2) \model\ performs better than BioLLMs and GPT-3.5, it indicates the effectiveness of local and external data sources.  
(3) Though the size of \model\ is much smaller than SciRAG (NewBing), it has better performance. The gain comes from two aspects. The first one is our customized prompts. The second aspect lies in the local and external information sources. NewBing has no access to specialized databases and lacks technical biological descriptions for reasoning.
(4) GeneGPT scores 0\% accuracy in this task, because it is a customized model for the GeneTuring dataset, resulting in poor generalization capabilities.

\begin{figure}[ht]
  \includegraphics[width=0.97\columnwidth]{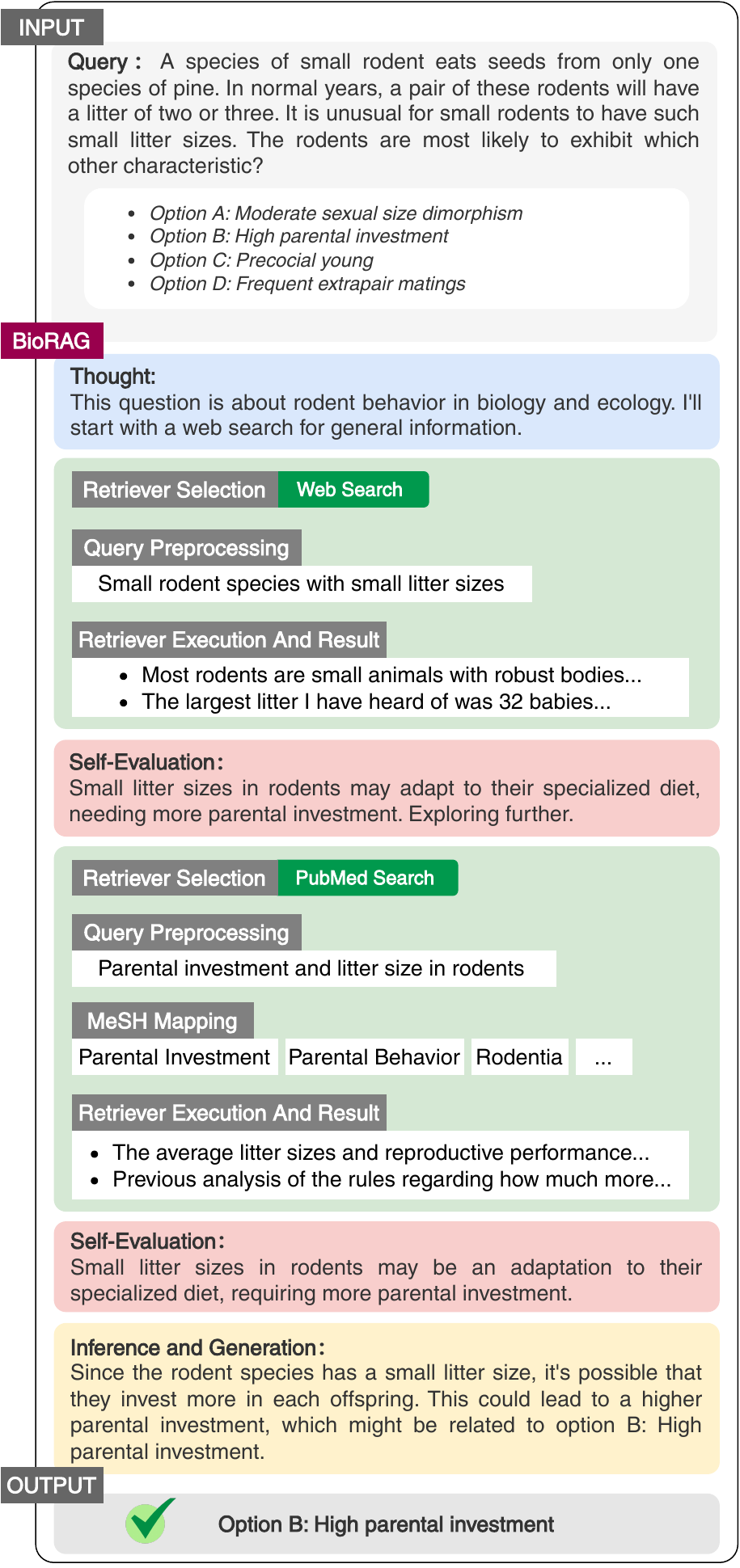}
  \caption{A case study selected from the College Biology dataset.}
  \label{fig:case_1}
\end{figure}

\begin{figure}[ht]
  \includegraphics[width=0.985\columnwidth]{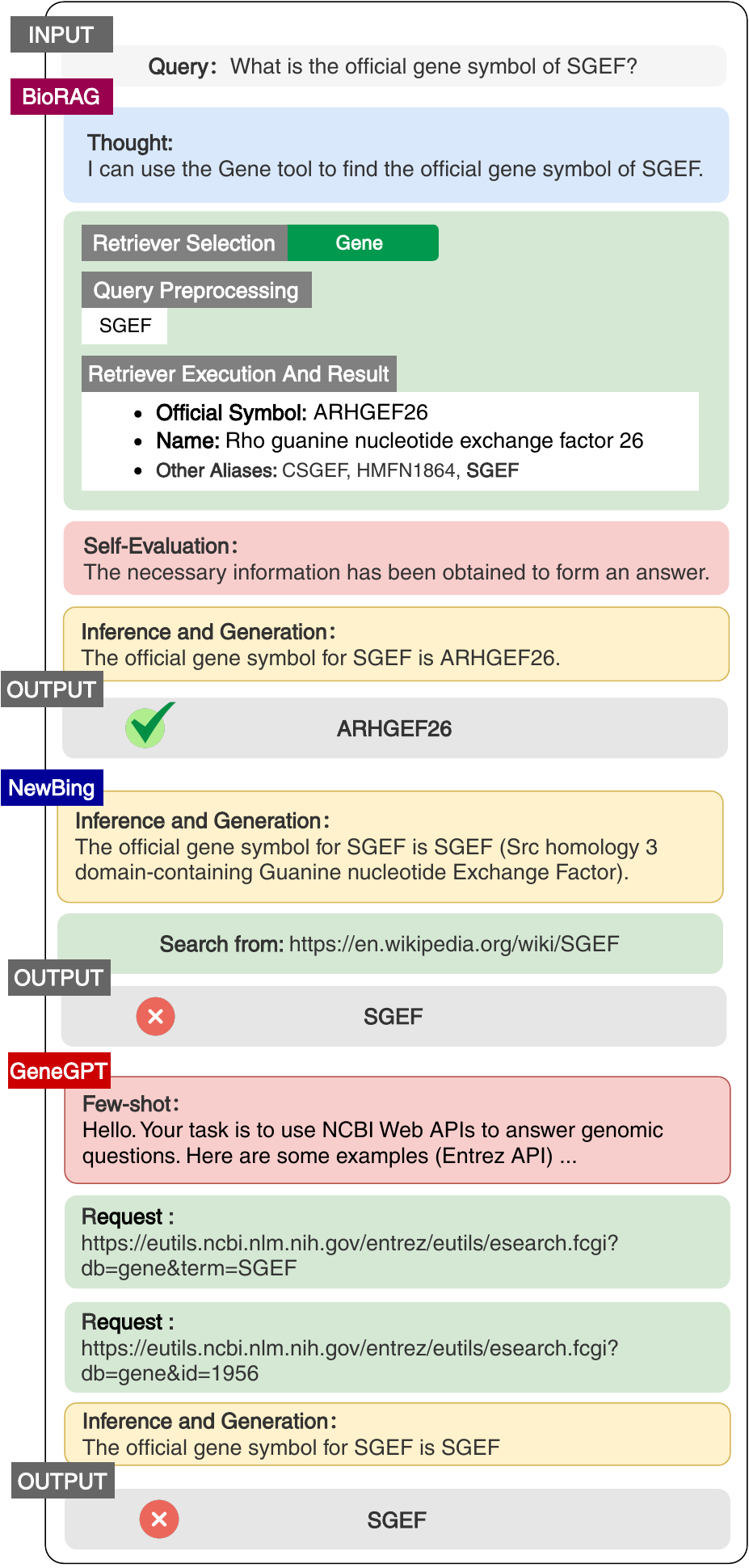}
  \caption{A case study conducted on the gene alias task in the GeneTuring dataset.}
  \label{fig:case_2}
\end{figure}

\begin{figure*}[htbp]
  \includegraphics[width=0.95\linewidth]{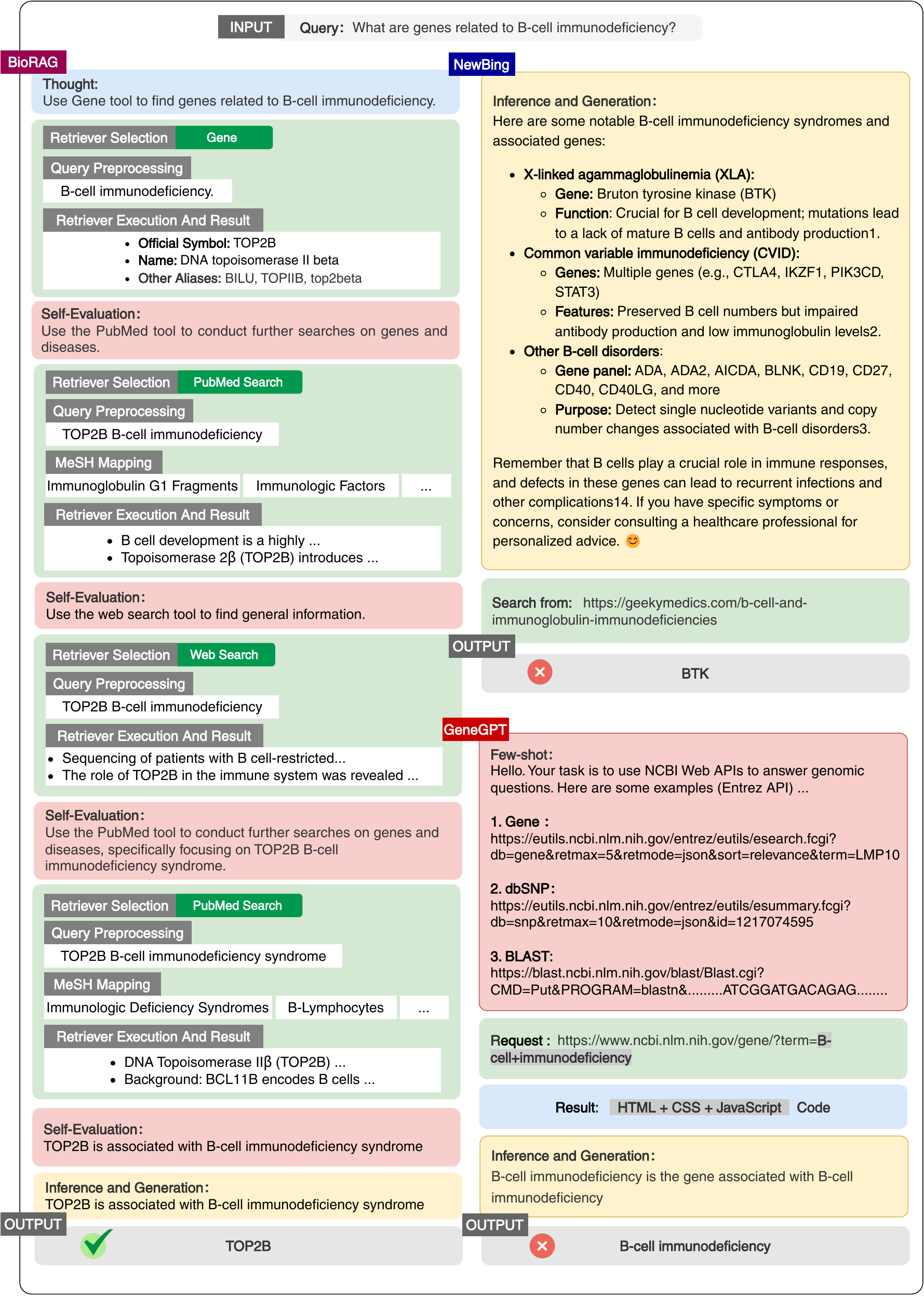}
  \caption{A case study conducted on the gene disease association task in GeneTuring dataset.}
  \label{fig:case_3}
\end{figure*}

\subsection{Specialized Biological Reasoning Results}
The GeneTuring dataset contains more specialized biological questions, and the corresponding reasoning process highly relies on technical biological corpus and descriptions. Results are shown in Table~\ref{tab:res_turing}. As this dataset does not contain the train data, BioLLMs are performed directly without finetuning. Their bad results indicate their poor generalization. In this dataset, we focus on the analyses of GeneGPT, NewBing, and \model\: (1) For the \textbf{nomenclature tasks}, the performance of \model\ and GeneGPT rank first and second respectively, as both of them have access to the Gene database. \model\ integrates the results of search engines while GeneGPT does not, and this brings the gap.
(2) The reasoning behind \textbf{genomic location tasks} relies on the highly specialized Gene and dbSNP database. \model\ and GeneGPT achieve 100\% accuracy in the gene SNP association sub-task, as both of them have access to the dbSNP database. However, NewBing has no access to the dbSNP database, thus it gets 0\% accuracy in this task. For the gene location subtask, the challenge is the variants of gene names. The interface of GeneGPT does not support advanced search, thus the retrieved names are not comprehensive. In contrast, general search engines, such as NewBing, have better retrieved results when the query entity has variants or ambiguities. Thus NewBing has a better performance in this task than GeneGPT. \model\ supports the above two kinds of interfaces, and achieves the best results in this task.
(3) \textbf{Functional analysis tasks} rely on both the Gene database and relative PubMed papers. The PubMed corpus provides detailed gene-disease relationships. Although NewBing retrieves the metadata, \model\ combines the local PubMed database with other specialized databases to achieve the best results.

\subsection{Ablation Study}
To evaluate the contribution of each component of \model, we performed an extensive ablation study using the GeneTuring dataset, systematically removing individual components to assess their impact on performance across various tasks. This study was designed to isolate the effects of different databases, components, and base models, with the experiments categorized as follows: (1) \textbf{Databases}: We consider three variations to evaluate the effectiveness of each data sources of our database: \textbf{D1}: \model without the Gene database; \textbf{D2}: \model without general search engines. \textbf{D3}: \model without the local PubMed database. (2) \textbf{Model Components}: We investigate the impact of specific components of our proposed framework: \textbf{C1}: \model without  the MeSH Filter; \textbf{C2}: \model without  the Query Rewrite component; \textbf{C3}: \model without  the Self-Evaluation mechanism. (3) \textbf{Base Models}: We compare the performance when using two different base LLM models:  \textbf{M1}: take Llama-3-8B as the basic LLM, and \textbf{M2}: take Llama-3-70B as the basic LLM of BioRAG.

Based on the results of ablation study, we highlights the following key findings: (1) \textbf{Impact of Databases:} The results indicate that the Gene database (D1) plays a crucial role in performance. For instance, the accuracy significantly drops in tasks such as  Gene\_location when this component is removed. The general search engines (D2) and local PubMed database (D3) also contribute positively, but their impact is less pronounced compared to the Gene database. (2) \textbf{Component Contributions:} Among the components, the Self-Evaluation mechanism (C3) is vital for maintaining high accuracy across most tasks. The MeSH Filter (C1) and Query Rewrite (C2) also enhance performance, but their absence does not degrade the results as severely as the removal of Self-Evaluation. (3)  \textbf{Effects of Basic Language Models:} Comparing the two base models, Llama-3-70B (M2) generally outperforms Llama-3-8B (M1) across all tasks, indicating that the larger model size contributes to better handling of complex biological queries. These findings underscore the importance of integrating diverse data sources and advanced components within the \model\ framework to achieve optimal performance in biological question reasoning tasks.  By understanding the contribution of each component, we can better optimize \model\ for different tasks and datasets.

\subsection{Case Study}
To compare reasoning differences among \model\ and the baselines in a more intuitive manner, we select three typical case studies in this section.
We first provide a case study to show the workflow of \model\ (Figure~\ref{fig:case_1}). It is selected from the College Biology dataset. \model\ performs self-evaluation twice: the first time it starts with a web search for general information, but the results are insufficient to support answering the question. Thus \model\ conducts the second self-evaluation and calls for the more specialized PubMed database. The results this time are accurate and sufficient to support answering the question, thus \model\ gives the final answer based on the results.

The second case study is conducted on the gene alias task in the GeneTuring dataset (Figure~\ref{fig:case_2}). The challenge of this task is the variants of gene names. NewBing gets the response from the Wikimedia. However, Wikimedia is not specialized enough to provide the alias for the input gene, which leads to the wrong answer. 
The prompts of GeneGPT are too complicated, none of the prompts is relevant to this task. In addition, its NCBI API returns the gene IDs, instead of the gene names. The LLM is unable to understand these IDs, and finally arrives at a wrong answer. \model\ employs fuzzy queries, yielding a larger number of related responses with a higher error tolerance. Furthermore, each result contains detailed gene-related information and descriptions, such as the aliases. Thus \model\ gets the correct answer.

The third case study is conducted on the gene-disease association task in the GeneTuring dataset, shown in Figure~\ref{fig:case_3}. Reasoning behind this task relies on both the Gene database and relative PubMed papers. The PubMed abstracts provide detailed gene-disease relationships.
NewBing gets the response from the Geekymedics website. Although the Geekymedics website provides general medical information, it does not offer the correct or specific details required for gene-disease associations. Consequently, NewBing's response is inaccurate due to the reliance on a non-specialized source.
GeneGPT chose the wrong NCBI API. The API's feedback is a complicated and interminable HTML page, with massive irrelevant information or descriptions. Based on the ambiguous backgrounds, GeneGPT outputs the wrong answer.
In the reasoning process of \model, BioRAG uses multiple tools, i.e., Gene database, local PubMed database, and Web search, to gather and conduct mutual confirmation on the information of genes associated with B-cell immunodeficiency. The process involves preprocessing queries, executing searches, and conducting self-evaluations at each step to ensure comprehensive and accurate results. The reasoning process is thorough, incorporating various data sources to confirm the association of specific genes with B-cell immunodeficiency.

\section{Conclusion}
This paper introduces \model, an innovative framework that integrates Retrieval-Augmented Generation with Large Language Models to enhance biological question-reasoning.
The framework's ability to obtain relevant and current information from a blend of traditional databases, toolkits, and modern search engines ensures the accuracy of the generated answers.
Through extensive validation, including rigorous testing on widely recognized biology QA datasets and extensive case studies, \model\ has demonstrated its superior ability to handle complex biological queries. These results underscore the framework's potential as a valuable tool for the scientific community, facilitating more accurate and efficient information processing.

\end{document}